\documentclass[conference]{IEEEtran}
\IEEEoverridecommandlockouts
\usepackage{cite}
\usepackage{array}
\usepackage{float}
\usepackage{amsmath,amssymb,amsfonts}
\usepackage{algorithm}
\usepackage{algpseudocode}
\usepackage{graphicx}
\usepackage{textcomp}
\usepackage{xcolor}
\usepackage{booktabs}
\def\BibTeX{{\rm B\kern-.05em{\sc i\kern-.025em b}\kern-.08em
    T\kern-.1667em\lower.7ex\hbox{E}\kern-.125emX}}
\begin{document}

\title{Large Models Enabled Ubiquitous Wireless Sensing\\
}

\author{\IEEEauthorblockN{Shun HU}
\IEEEauthorblockA{\textit{School of Sicence and Engineering} \\
\textit{The Chinese University of Hong Kong (Shen Zhen)}\\
Shen Zhen, China \\
223010027@link.cuhk.edu.cn}
}

\maketitle

\begin{abstract}
In the era of 5G communication, the knowledge of channel state information (CSI) is crucial for enhancing network performance. This paper explores the utilization of language models for spatial CSI prediction within MIMO-OFDM systems. We begin by outlining the significance of accurate CSI in enabling advanced functionalities such as adaptive modulation. We review existing methodologies for CSI estimation, emphasizing the shift from traditional to data-driven approaches. Then a novel framework for spatial CSI prediction using realistic environment information is proposed, and experimental results demonstrate the effectiveness. This research paves way for innovative strategies in managing wireless networks.
\end{abstract}


\section{Introduction}
In the background of widespread 5G, the users around the globe benefit from high-speed characteristics of such a new communication technology. Meanwhile, the demand for more reliable and low-latency transmission is also increasing. Efficient management and adjustment of large-scale antenna arrays is an essential key to optimizing communication. Traditional management techniques are typically reactive, relying on feedback from users to adjust to the constantly changing wireless environment. But recent research emphasizes a forward-looking approach using CSI prediction. This proactive prediction-based management allows for more dynamic adjustment of network resources. \cite{zhang2024}

To explain the importance of CSI, it is needed to point out that CSI is essential for advanced functionalities of communication system such as precoding, bit-loading, adaptive modulation, channel-aware scheduling, and beamforming. In a MIMO setup, having precise knowledge of the CSI at the transmitter enables significantly higher channel capacity compared to transmission without CSI, improving the reliability of traditional communication systems. \cite{al-asadi2023}

In this context, CSI prediction is becoming an integral part. People used to utilize different approaches to predict channels, and various parametric models have been explored, such as the autoregressive (AR) model, the sum-of-sinusoids model, and the polynomial extrapolation model. As for concrete computation, a prediction algorithm called Prony-based angular-delay domain (PAD) was introduced as well. Despite their usefulness, these methods may struggle to accurately capture the complex features of real-world channels.

In recent years, researchers have sought innovative techniques to tackle the complexity of CSI prediction, ranging from traditional algorithms to modern algorithms, so the technique has evolved significantly. Meanwhile, AI tools are growing at a high pace, which helps in increasing the precision of prediction.

The content of this report includes:
\begin{itemize}
    \item Foundational Concepts in Wireless Channels: An overview of wireless channel theory, channel modeling, Massive MIMO, and OFDM fundamentals, establishing a basis for advanced prediction techniques.
    \item Time-Series Channel Prediction with \textit{LLM4CP}: Replication of time-series prediction experiments using the “LLM4CP” framework.
    \item Spatial CSI Prediction Studies: Exploration of spatial CSI prediction with large models.
\end{itemize}

The contributions of this paper include:
\begin{itemize}
    \item Application of Large models for Spatial CSI Prediction.
    \item Development of a preprocessing framework which merges geometric-physical knowledge for robust prediction.
\end{itemize}

The notations which could be used are presented in the table below.\\

\hspace{0.8cm}
\begin{tabular}{| l || l |} 
    \hline 
    $\mathbf{A}_{p \times q}$ & $p \times q$ matrix \\ \hline
    $||\mathbf{A}||_2$ & Frobenius norm of matrix \\ \hline
    $\mathbf{A}^{T}$ & transpose of $\mathbf{A}$ \\ \hline
    $\mathbf{A}^{H}$ & Hermitian of $\mathbf{A}$ \\ \hline
\end{tabular}

~\\
\section{Review on Related Works}
The foundation of modern CSI prediction can be traced back to early works like \cite{gheryani2009}, which addressed the challenge of adaptive channel estimation in MIMO systems, focusing on the use of feedback channels to optimize transmission. The work by \cite{abd2018} highlights the challenges in traditional methods of estimating channel coefficients. 

Building on this, \cite{liu2019} address the limitations imposed by the dimensionality of CSI in massive MIMO systems. They propose leveraging compressive sensing and deep learning to enhance the efficiency of CSI estimation, as a groundbreaking work.

In 2020, \cite{kim2020} further advanced the discussion by comparing traditional Kalman filtering techniques with emerging machine learning methods. Then the research shifts towards data-driven methodologies is echoed in the work of \cite{pecorella2020}, who thoroughly explore the potential of deep learning to enhance CSI prediction accuracy.

\cite{zhang2022} contribute to this discourse by proposing a novel 3D CNN framework for predicting future CSI in highly mobile environments. This trend towards utilizing advanced neural network architectures is further supported by \cite{ko2022}, who introduce a domain transformation technique to enhance training efficiency for channel prediction, highlighting the ongoing refinement of machine learning applications in CSI estimation.

The exploration of meta-learning and deep denoising techniques by \cite{kim2022} illustrates how such new strategies can alleviate the challenges posed by limited pilot data, thereby enhancing the performance of channel predictors in real-world scenarios. Apart from that, \cite{zhou2022} expand the range of channel prediction to spatial-temporal domain, where two different tools CNN and convolutional variation of LSTM is combined to be namely Conv-LSTM to capture the inherent complexity of massive MIMO systems, which can be found in \cite{kim2022}.

In a more recent study, \cite{du2023} delve into the applicability of meta-learning for adaptive predictors to new environments, emphasizing the necessity for models that can quickly adjust to varying conditions without extensive retraining. 

In the context of advancing 6G technologies, \cite{zhang2024} introduces a novel framework, the digital radio twin, which leverages U-Net for accurate spatial-CSI prediction.

Finally, \cite{zhou2024} provides a comprehensive survey on LLMs in telecommunications, highlighting their capabilities in managing complex network tasks and optimizing CSI prediction processes.
~\\

\section{Basic Concepts of Channels}

\subsection{General Theory of Channel Models}
In wireless communication systems, the channel is a critical factor in determining the overall performance. In a MIMO system with \( m \) transmitting antennas and \( n \) receiving antennas, the channel is typically represented by a matrix, named CSI, as shown below:

\begin{equation}
H(t,\tau) = 
\begin{bmatrix}
h_{11}(t,\tau) & h_{12}(t,\tau) & \ldots & h_{n1}(t,\tau) \\
h_{21}(t,\tau) & h_{22}(t,\tau) & \ldots & h_{n2}(t,\tau) \\
\vdots       & \vdots      & \ddots & \vdots \\
h_{1m}(t,\tau) & h_{2m}(t,\tau) & \ldots & h_{nm}(t,\tau)
\end{bmatrix}
\end{equation}

The models of channels can be categorized in several ways, and the  diagram is shown in Fig.1, which shows three ways of classification: physical models, analytical models, and standard models.

\begin{figure}
    \centering
    \includegraphics[width=1\linewidth]{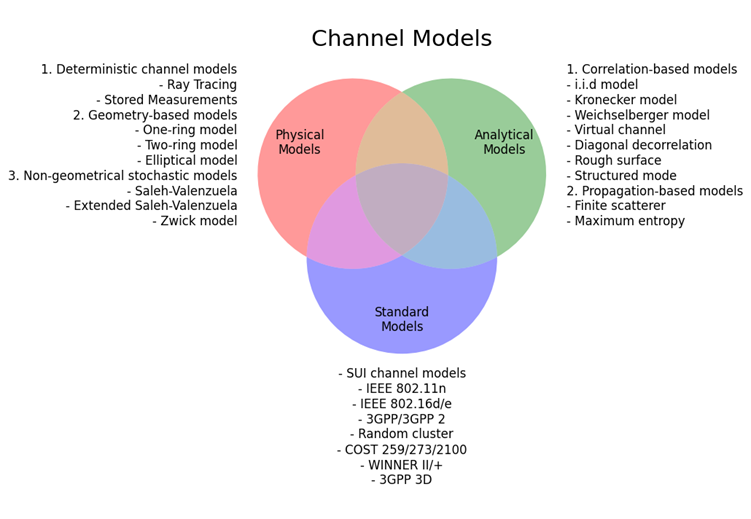}
    \caption{Modern classification of wireless channels}
    \label{fig:enter-label}
\end{figure}

Physical models use basic theories in physics to describe the multipath propagation environment between the transmitter and receiver, considering factors like the direction of arrival (DoA), direction of departure (DoD), signal amplitude, and the delay of multipath components (MPCs). 

Analytical models focus on the channel impulse response or transfer function between specific pairs of transmitter and receiver antennas. These models do not directly account for wave propagation but instead synthesize a channel gain matrix that describes the overall system behavior. 

Standard models are designed mostly by industrial organizations and companies to assist in new wireless communication technologies. 

\subsection{OFDM Systems}
In modern digital wireless communication systems, orthogonal frequency division multiplexing (OFDM) technology is widely used to combat frequency-selective fading in multipath propagation environments, such as in standard IEEE 802.11a/g/n.

\section{WINNER Models and Ray Tracing Models}
Some of the channel dataset used in this work is generated by QuaDRiGa, which is based on the earlier WINNER model. Other data are generated using ray-tracing models, for realistic modelling and more stable prediction.

\subsection{WINNER}
WINNER supports a wide frequency range, making it suitable for mmWave and for scenarios with diverse transceiver techniques. WINNER models are adopted by QuaDRiGa. \cite{winner2008}.

\subsection{Ray Tracing}

Another important channel modeling used in this study is based on ray tracing, which is a deterministic method used to model wave propagation by simulating the paths of electromagnetic waves through an environment. Unlike statistical models that provide average estimates, ray tracing computes the interactions between waves and obstacles using the principles of geometric optics. This results in a more precise description of the channel characteristics.

\subsection{Softwares Using Ray Tracing}

WinProp (Wireless In-building Propagation) integrates a range of deterministic, empirical, and hybrid models, making it well-suited for wireless network planning and coverage analysis. Winprop’s interests go further than just principle-level verification since it can also be used to generate realistic datasets. WinProp supports the import of CAD files, offering flexibility in modeling different environments. 

In practice, One can adopt different approaches based on the electrical construction size and complexity (Fig.2). It is reasonable that in large urban scenarios, more approximations will be introduced in simulation, instead of solving complex wave equations. 
\begin{figure}
    \centering
    \includegraphics[width=1\linewidth]{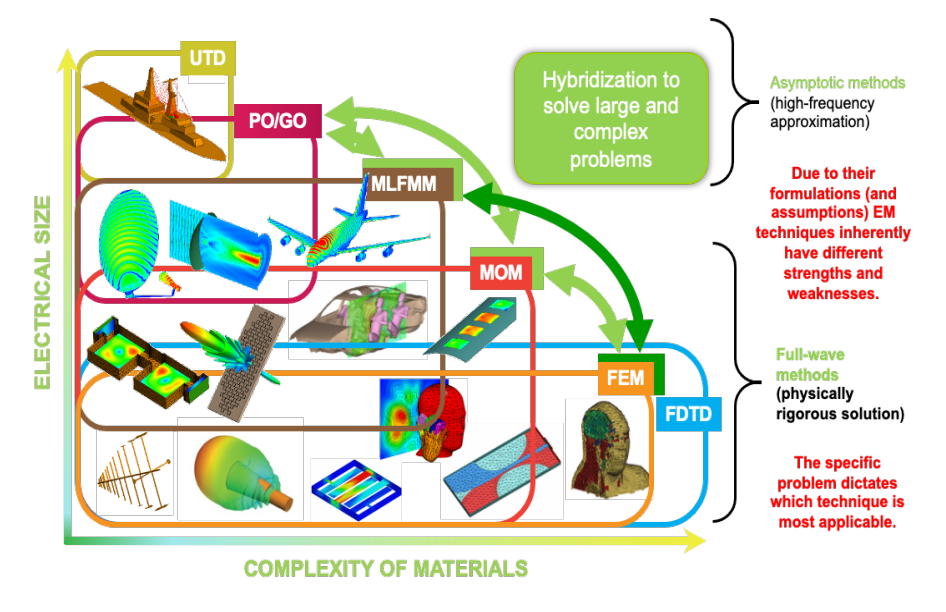}
    \caption{Numerical Methods of Winprop, with more approximation introduced when the prolem's scale becomes larger.}
    \label{fig:enter-label}
\end{figure}

\section{Reproducing Time-Series Channel Prediction Experiment}
As part of research, we will review and validate one related work in the field of intelligent channel prediction. The experiment in the paper \textit{LLM4CP: Adapting Large Language Models for Channel Prediction} (in short “\textit{LLM4CP}”) will be reproduced, where a pre-trained LLM is applied to predict future downlink CSI based on historical uplink CSI. 

\subsection{Modelling of Channel in \textit{LLM4CP}}
The base station is modeled with a dual-polarized uniform planar array (UPA), consisting of $N_t=N_h\times N_v$ antennas. The users are set to have motion, which induces Doppler shifts and contributes to the time variation of the channel. 

The received signal at UE is given by:

\begin{equation}
y_k = h_k^H w_k x_k + n_k,
\end{equation}

where $h_k$ is the downlink CSI for the $k$-th subcarrier, $w_k$ is the transmit precoder, $x_k$ is the transmitted symbol, and $n_k$ is additive white Gaussian noise (AWGN). The precoding vector $w_k$ is designed based on $h_k$, and any inaccuracy in the predicted CSI can result in suboptimal transmission rates.

\subsection{Formulation of Channel Prediction Problem}

In \textit{LLM4CP}, the goal is to predict future downlink CSI over $ L$ RBs using historical uplink CSI from $P$ RBs. In mathematical formulation, we hope to minimize NMSE:

\begin{equation}
\text{NMSE} = \mathbb{E} \left\{ \frac{\sum_{i=1}^{L} \| \hat{\mathbf{H}}_d^{S+i} - \mathbf{H}_d^{S+i} \|_F^2}{\sum_{i=1}^{L} \| \mathbf{H}_d^{S+i} \|_F^2} \right\}
\end{equation}

Where:
\begin{itemize}
    \item $\mathbb{E}$ represents the expectation over all the data samples.
    \item $\hat{\mathbf{H}}_d^{S+i}$ represents the predicted downlink CSI at time \(S+i\).
    \item $\mathbf{H}_d^{S+i}$ represents the actual downlink CSI at time \(S+i\).
\end{itemize}

The uplink and downlink CSI are assumed to share some statistical correlation due to channel reciprocity, which can be leveraged to infer future downlink conditions.

\subsection{Experiment Methodology}
\subsubsection{CSI Generation}
The first step of the experiment is to generate channel data. To replicate real-world communication environments, we configure standardized 3GPP 38.901 Urban Macro(Uma) scenario using QuaDRiGa. Users are set to move linearly at speeds ranging from 10 km/h to 100 km/h. 

\begin{table}[h]
    \centering
    \caption{Parameters Configuration}
    \begin{tabular}{|l|l|}
        \hline
        \textbf{Parameter} & \textbf{Value} \\
        \hline
        Center Frequency & 2.4 GHz \\
        BS Antenna & 4x4 dual-polarized array \\
        Antenna Tilting Angle & 7 degrees \\
        User Device & Single omni-directional antenna \\
        User Speed & 10.1\~100 km/h in 0.1 km/h increments \\
        Velocity Sampling Period & 0.5 ms \\
        Sample Duration & 19 cycle lengths \\
        \hline
    \end{tabular}
    \label{tab:parameters_configuration}
\end{table}

In order to have a direct comprehension of the experimental setup, the locations of UE and BS are depicted in Fig.3. The height of BS is 30m and the height of UEs is 0m.

\begin{figure}
    \centering
    \includegraphics[width=0.8\linewidth]{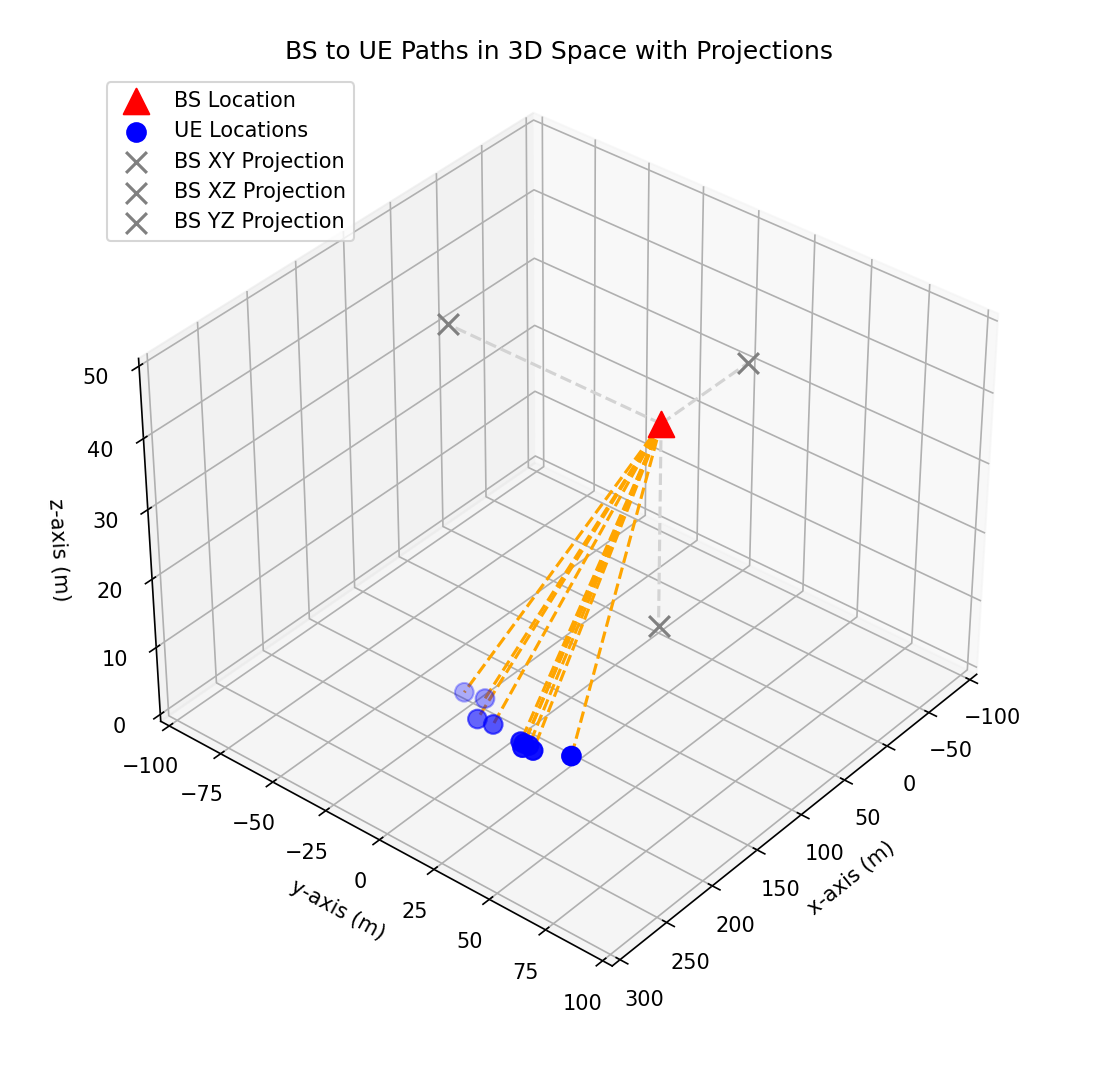}
    \caption{Locations of UEs and BS in \textit{LLM4CP}'s experiment setup. For consideration of simplicity, only LOS paths are shown. But remember that NLOS paths also exist and they are caused by randomly generated scatter points in the space, under QuaDriGa built-in algorithms.}
    \label{fig:enter-label}
\end{figure}

\subsubsection{CSI Data Preprocessing}
The complex channel matrix $\mathbf{H}_f$ is constructed for each transmit antenna $j$ using the temporal length of historical CSI data. 

\subsubsection{Model Training}

Both frequency domain and delay domain information are feeded into the model. Note that most layers of the pre-trained GPT-2 are frozen during training, while other parameters of the network are trainable. The number of trainable parameters is relatively small.

\begin{table*}
    \centering
    \begin{tabular}{lcccccccccc}
        \toprule
        Model/Speed(km/h) & 10 & 20 & 30 & 40 & 50 & 60 & 70 & 80 & 90 & 100 \\
        \midrule
        GPT(LLM4CP)         & 0.02083 & 0.02296 & 0.02485 & 0.02608 & 0.02983 & 0.03292 & 0.03849 & 0.04231 & 0.04986 & 0.05747 \\
        Transformer & 0.03138 & 0.03286 & 0.03429 & 0.03232 & 0.04195 & 0.04371 & 0.05647 & 0.06176 & 0.07890 & 0.08816 \\
        CNN         & 0.02882 & 0.03075 & 0.03101 & 0.03113 & 0.03823 & 0.04109 & 0.05071 & 0.05773 & 0.07105 & 0.08244 \\
        GRU         & 0.05893 & 0.06010 & 0.05556 & 0.04230 & 0.06078 & 0.05665 & 0.08096 & 0.08246 & 0.10270 & 0.09824 \\
        LSTM        & 0.07976 & 0.07787 & 0.06596 & 0.04831 & 0.07003 & 0.06375 & 0.08819 & 0.09591 & 0.12286 & 0.11560 \\
        RNN         & 0.16996 & 0.13275 & 0.10381 & 0.07724 & 0.10493 & 0.08652 & 0.11641 & 0.12414 & 0.15519 & 0.20657 \\
        No Prediction          & 0.05873 & 0.13305 & 0.25106 & 0.41127 & 0.59231 & 0.79652 & 1.02281 & 1.25426 & 1.47749 & 1.68876 \\
        PAD         & 0.08252 & 0.12101 & 0.14623 & 0.11190 & 0.17296 & 0.18224 & 0.23769 & 0.15349 & 0.18944 & 0.18723 \\
        \bottomrule
    \end{tabular}
    
    \vspace{0.5em}
    \caption{NMSE of prediction for different models across speeds, under TDD scenario}
    \label{tab:nmse_values}
\end{table*}

\begin{table*}
    \centering
    \begin{tabular}{lcccccccccc}
        \toprule
        Model/Speed(km/h) & 10 & 20 & 30 & 40 & 50 & 60 & 70 & 80 & 90 & 100 \\
        \midrule
        GPT(LLM4CP)         & 0.40161 & 0.34878 & 0.35639 & 0.40715 & 0.34129 & 0.38660 & 0.40247 & 0.33781 & 0.36739 & 0.40094 \\
        Transformer & 0.56755 & 0.49624 & 0.47638 & 0.54351 & 0.48867 & 0.52255 & 0.51155 & 0.48147 & 0.49504 & 0.51693 \\
        CNN         & 0.97241 & 0.96062 & 0.96784 & 0.95742 & 0.95547 & 0.94933 & 0.94312 & 0.97714 & 0.96508 & 0.97834 \\
        GRU         & 0.62848 & 0.60466 & 0.60371 & 0.61128 & 0.59004 & 0.59892 & 0.59454 & 0.60476 & 0.60998 & 0.61508 \\
        LSTM        & 0.69523 & 0.65747 & 0.64816 & 0.65555 & 0.62290 & 0.64631 & 0.63110 & 0.64763 & 0.64663 & 0.66668 \\
        RNN         & 0.60296 & 0.55399 & 0.53408 & 0.52743 & 0.51272 & 0.50512 & 0.50732 & 0.52599 & 0.53564 & 0.58028 \\
        No Prediction          & 2.01695 & 1.92730 & 1.95513 & 1.93453 & 1.99573 & 1.88920 & 1.92420 & 1.85007 & 1.89675 & 1.82435 \\
        \bottomrule
    \end{tabular}
    
    \vspace{0.5em}
    \caption{NMSE of prediction for different models across speeds, under FDD scenario}
    \label{tab:old_nmse_values}
\end{table*}

\subsubsection{Model Evaluation}
To evaluate the performance of the proposed LLM4CP model, we compare its NMSE with several baselines across different user velocities for both TDD and FDD systems, and these include PAD, RNN, LSTM, GRU (Gate recurrent unit), CNN and traditional Transformers (Fig.4 and Fig.5).

\begin{figure}
    \centering
    \includegraphics[width=1\linewidth]{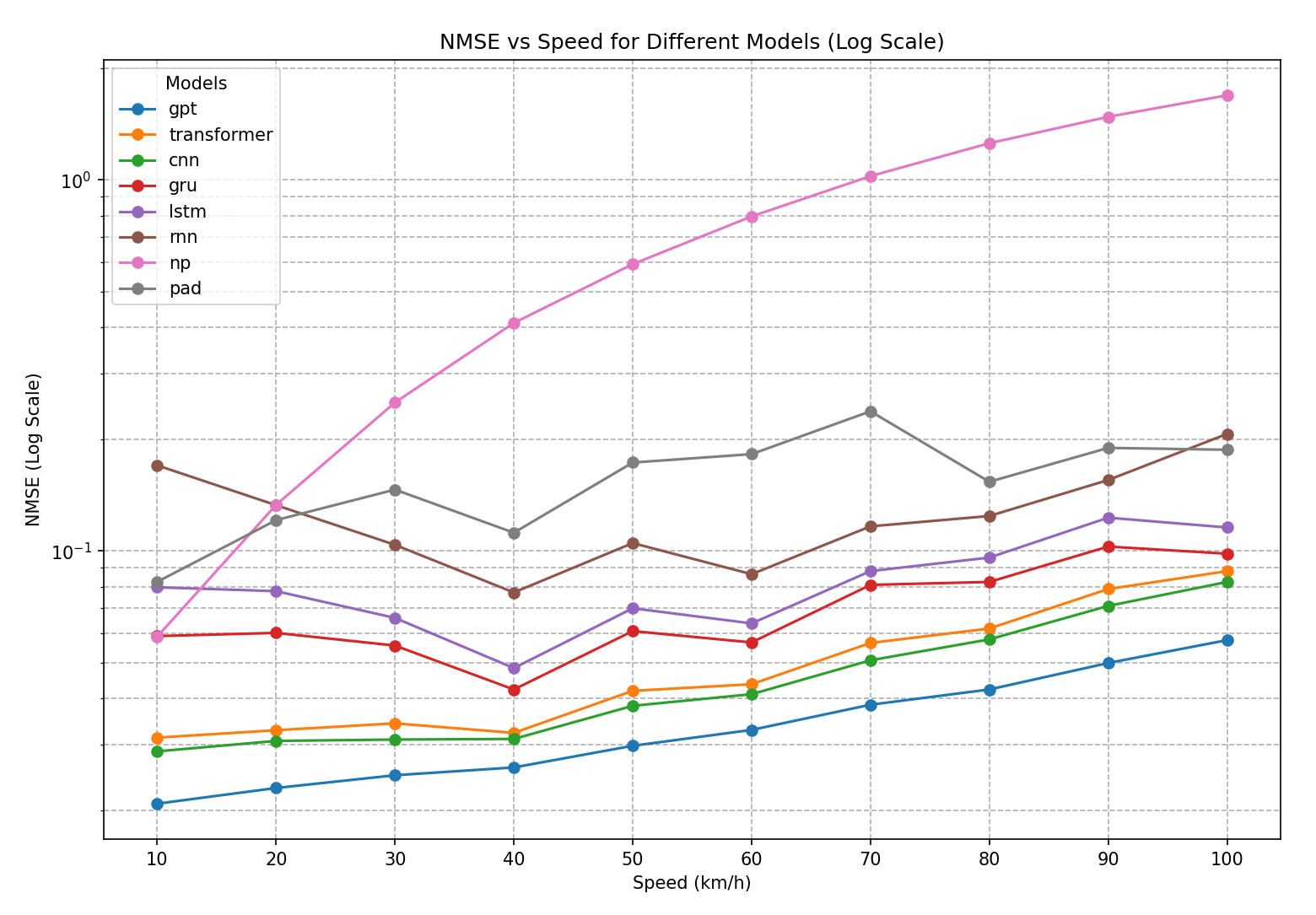}
    \caption{The LLM4CP and other baselines' NMSE performance in relation to various user velocities for TDD systems}
    \label{fig:enter-label}
\end{figure}

\begin{figure}
    \centering
    \includegraphics[width=1\linewidth]{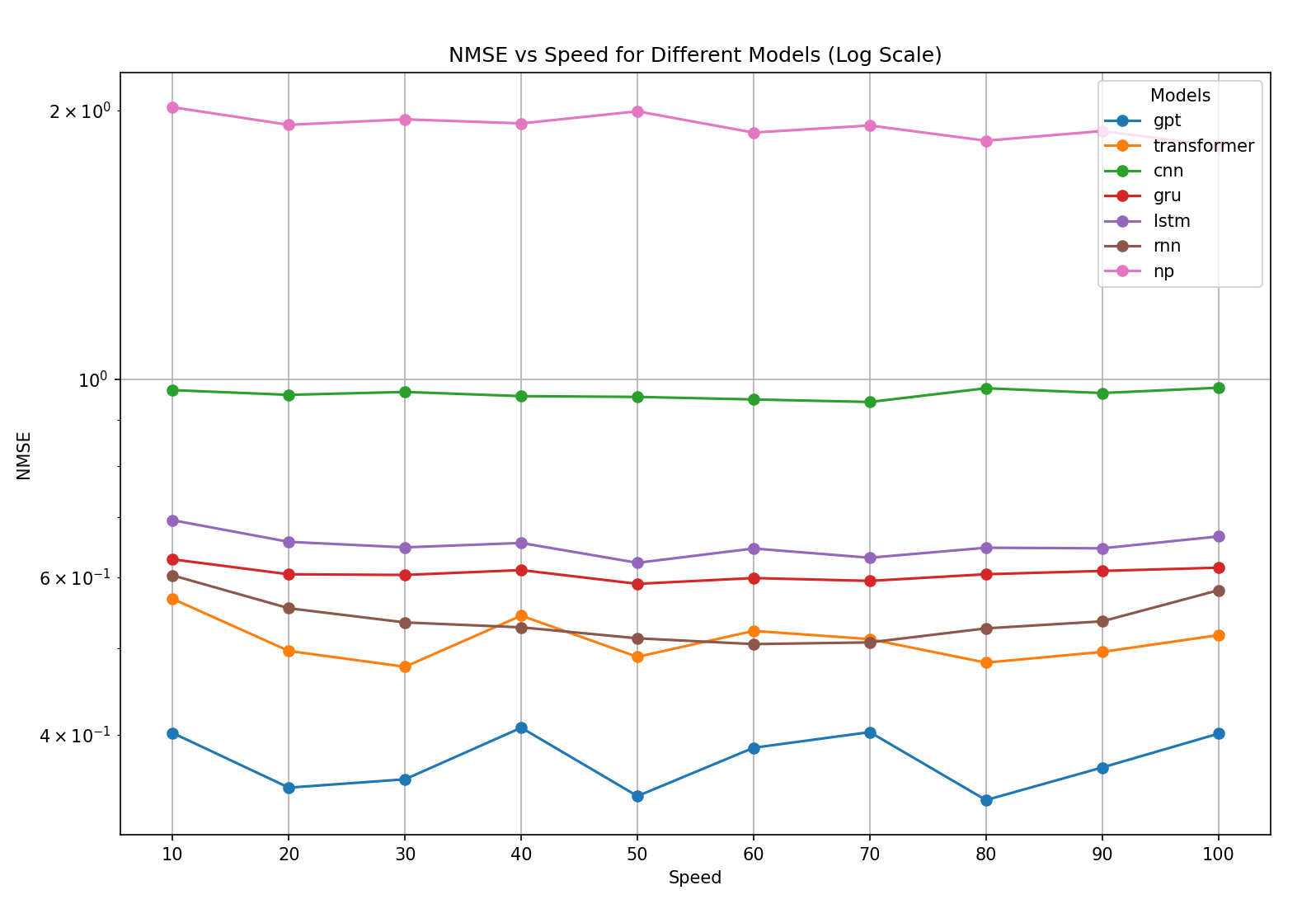}
    \caption{The LLM4CP and other baselines' NMSE performance in relation to various user velocities for FDD systems}
    \label{fig:enter-label}
\end{figure}

\section{QuaDRiGa Dataset for Spatial CSI Prediction}

Our work will be focused on CSI spatial prediction from now on. This section serves as a heuristic startup and tools will be refined in later sections. Spatial prediction could be beneficial when the BS acquires knowledge about a user's location but communication rate between BS and this user is relatively slow.

We still adopt QuaDRiGa for data generation, with the main adjustment being that all pedestrians are stationary to focus on spatial rather than temporal prediction. Users are positioned within a 3D area (horizontal range: -100 m to 100 m, height: 0 to 3 m).

\subsection{Training Procedure}

Model training employs a customized GPT-2-based structure, with AdamW optimizer. During training, the input to the model is each position $\mathbf{p} \in \mathbb{R}^3$, and the output is the predicted CSI value $\mathbf{H}^{de}$ for the corresponding position. The predicted CSI $\hat{\mathbf{H}}$ is hoped to be as close as possible to the ground-truth CSI $\mathbf{H}^\text{gt}$.

The training process uses a smooth L1 loss function, $\mathcal{L}_{\text{SmoothL1}}$. This loss function is a combination of absolute error and squared error:

\begin{equation}
\mathcal{L}_{\text{SmoothL1}} = 
\begin{cases} 
0.5 ||\mathbf{H}^{\text{de}} - \mathbf{H}^\text{gt}||^2, & \text{if } || \mathbf{H}^{\text{de}} - \mathbf{H}^\text{gt} ||_F^2 < 1 \\
|| \mathbf{H}^{\text{de}} - \mathbf{H}^\text{gt} ||_F^2 - 0.5, & \text{otherwise}
\end{cases}
\end{equation}

\subsection{Results of Spatial CSI Prediction Based on QuaDRiGa}

The model incorporates dual convolutional layers before GPT-2. Data from single time stamp leads to the lowest observed validation NMSE of 0.17. Then, model is trained using multiple time slices. This leads to a noticeable improvement in performance and the least NMSE is 0.035.

\subsection{Limitations of Current Approach and Improvements}

While the current work demonstrates seemly ideal results, several limitations still need to be addressed. The greatest drawback is that, current implementation uses a statistical channel. The scatter points in space are random, resulting in fluctuating channel conditions that do not correspond to physical existence. This randomness can adversely affect the stability and accuracy of predictions. 

\section{MIMO-OFDM Channel Modeling in WinProp for Realistic Spatial CSI Prediction}

\subsection{Methodology of Channel Modelling}
To achieve a realistic channel that accounts for surroundings and data frames in real-world settings, and realize high-level customization, we utilize WinProp. 

The whole procedure from channel simulation to machine learning is depicted in Fig.6.

\begin{figure}
    \centering
    \includegraphics[width=0.8\linewidth]{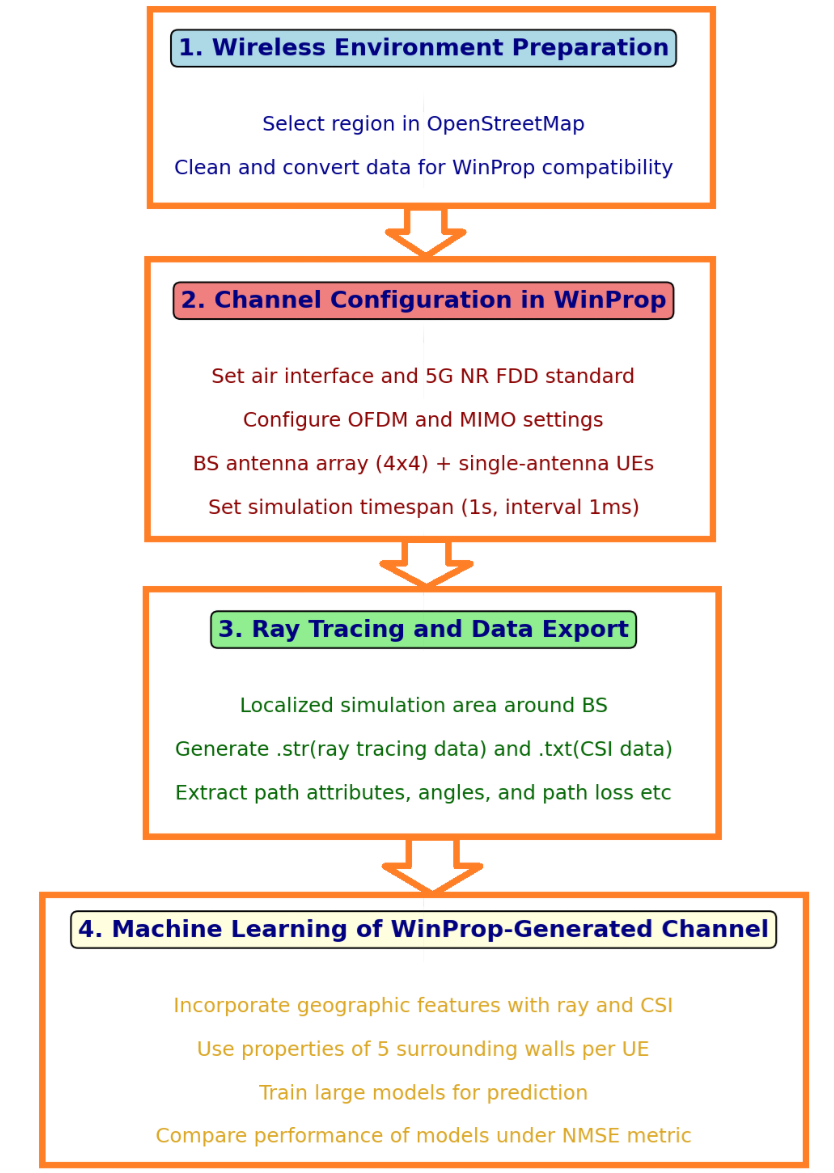}
    \caption{Procedure from physical environment buil-up, channel generation, to model training}
    \label{fig:enter-label}
\end{figure}

The first step involves preparing an environmental layout. We select the area of CUHK(SZ) in OpenStreetMap to obtain an OSM file, then clean and convert the data via QGIS for compatibility with WinProp, shown in Fig.7. The OSM file provides a detailed model of buildings, streets, and vegetation zones. Once imported, the simulation can be conducted in the area defined by the OSM file.
\begin{figure}
    \centering
    \includegraphics[width=0.7\linewidth]{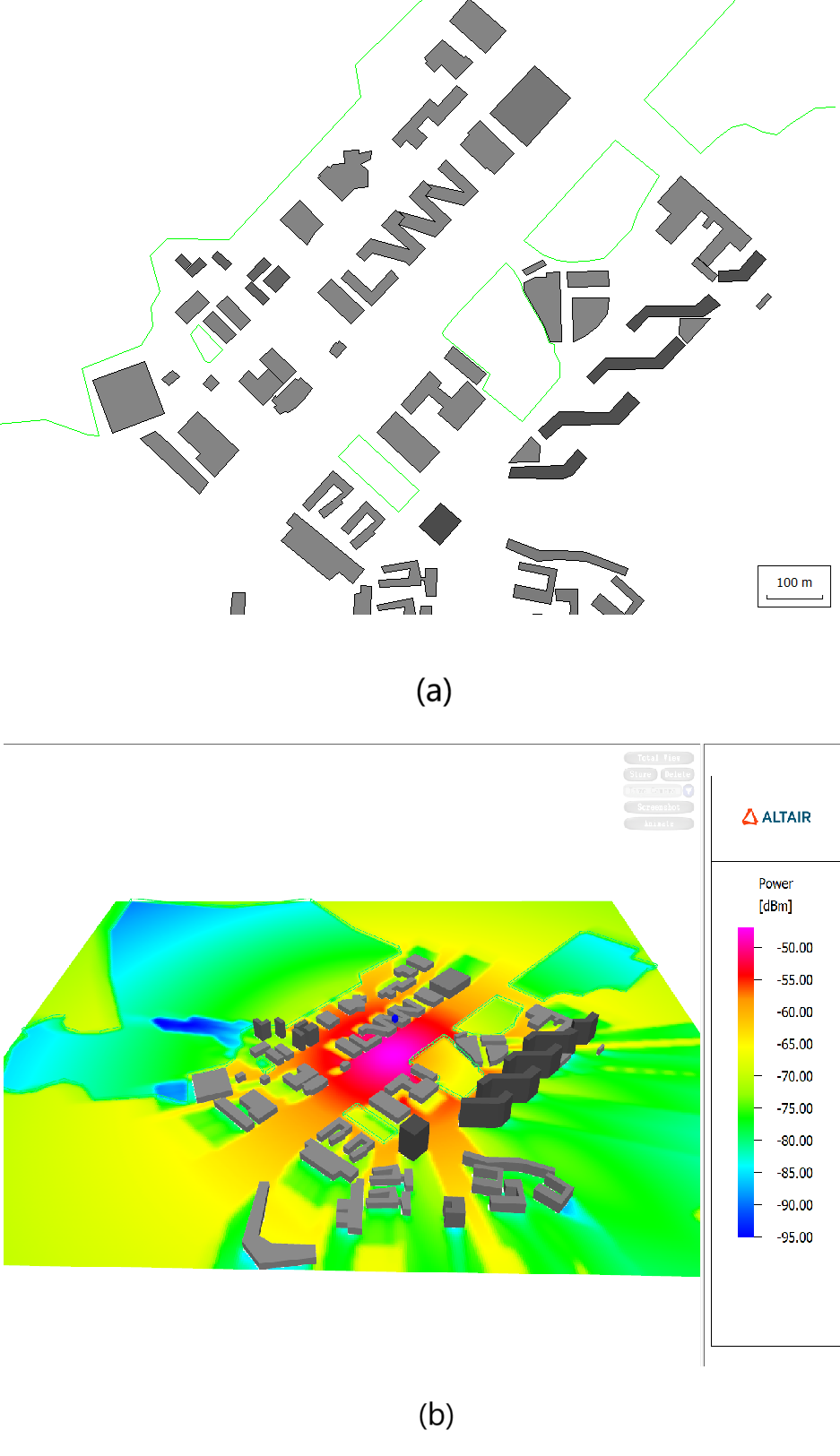}
    \caption{(a) The selected region of CUHK(SZ) as the place for ray-tracing simulation. (b) The wide range simulation result shown in power distribution, with a tiny blue dot at the center representing the BS. Notice that the simulation in the picture is coarse with distant UEs. For model training, more dense UEs are simulated in a smaller region.}
    \label{fig:enter-label}
\end{figure}
Then the OFDM protocol is selected in air interface settings, and the 5G NR FDD standard description file is imported to WinProp for consistency with current wireless communication protocols. It is possible to custom-tailor parameters such as subcarrier spacing, modulation schemes, and bandwidth of the system. For industrial-grade simulators like WinProp, without configuration the air interface, the BS will only send simple modulated waves, and in this case discussing about data frame and RBs etc. is impossible.

The OFDM symbols per slot are set to 14, ensuring the simulation closely reflects high-capacity, high-frequency communication. For MIMO configurations, we implement a 4x4 BS antenna array combined with single-antenna UE.

In WinProp simulation fashion, each UE's position is uniformly distributed in predefined zones, instead of randomly generated according to a probabilistic function. The distance between UEs in the grid is 0.1m. Meanwhile the BS antenna is placed near the center.

\subsection{Ray Tracing and Data Export}

Ray tracing is conducted in a rectangle inside the map. Although it is allowed to simulate the channel for only one time-slot in WinProp, it will bring a static result, which makes CSI output impossible. To obtain CSI of the channel, the dynamic simulation is carried out, with the time span of simulation at 1s and the time interval being 1ms.

Each simulation run generates essential parameters for machine learning integration. The .str file encapsulates the attributes of each path, such as loss of path, along with the UEs interacted, but it is not measurable in practice, thus deserted. The .txt file contains CSI values for all locations and time-stamps considered in the simulation.

\subsection{Machine Learning of WinProp-Generated Channel for CSI Prediction}
Once the geographic features have been extracted, they need to be merged with CSI data to train models, corresponding to the step 4 of Fig.6. To achieve this, features require suitable representation. If a model predicts well across different points or even across regions, it means that the model can infer the CSI accurately from the geographic information which can be practically measured.

In our initial approach, the walls nearby each UE is treated as a segment characterized by distance and orientation relative to the UE. With such two geometric parameters, it is believed that the surrounding of UE can be grasped by the large model, and other material-related parameters will be learned indirectly throughout the epochs. For every UE, the properties of top 5 nearest walls are extracted, which has computational efficiency but lacks of details. 

To overcome insufficient representation of features, a new approach is adopted. As shown in Fig.8, the first stage involves embedding TX and RX positions into a raw geographical map and extracting environmental features. The map is then processed through a CNN module. The features are processed through multiple convolution and pooling layers to capture hierarchical representations of the environment. Intermediate features are concatenated to preserve multiscale information. 

\begin{figure}
    \centering
    \includegraphics[width=1\linewidth]{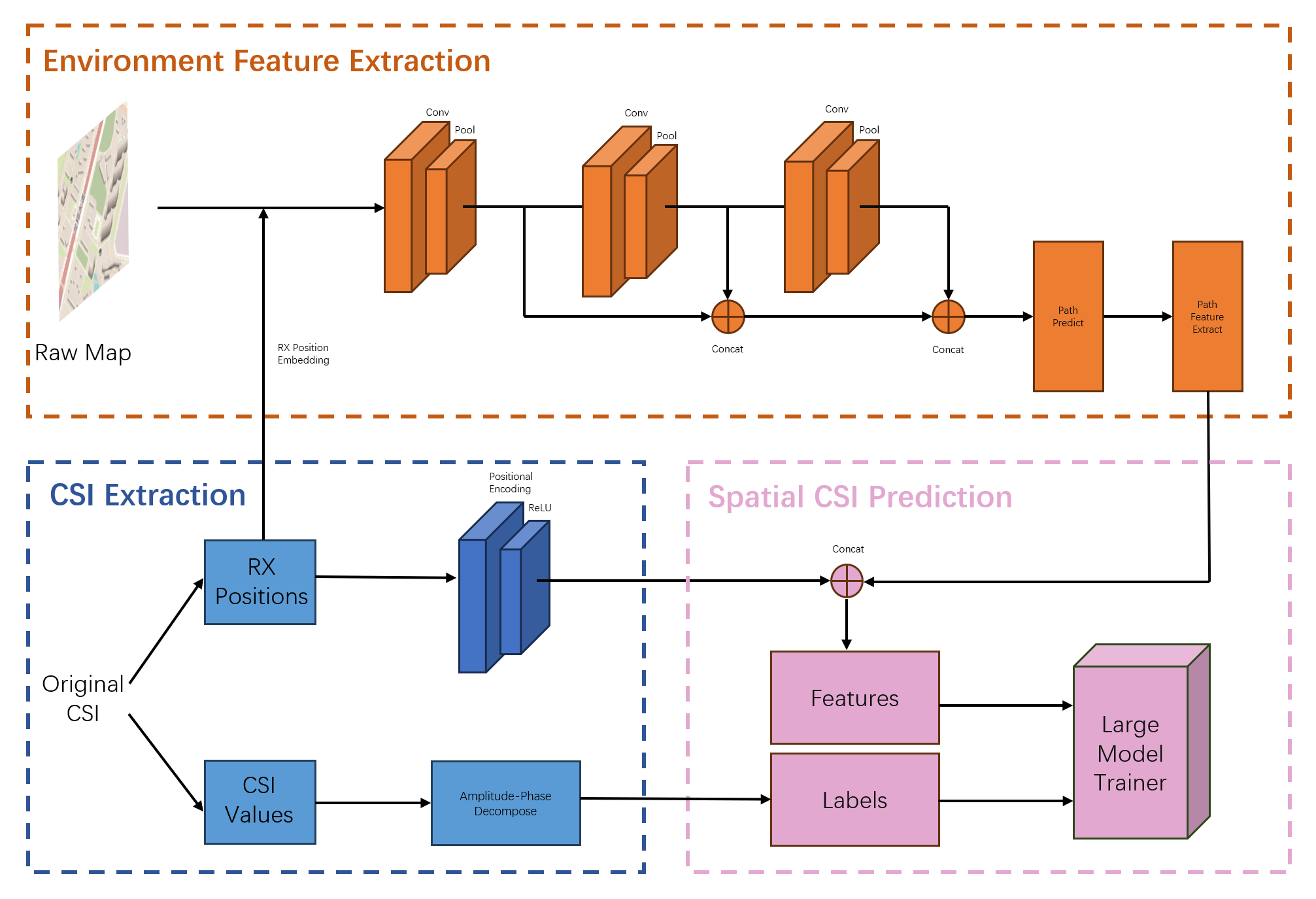}
    \caption{Novel architecture of machine learning targeted at informing CSI accurately from environmental knowledge, with prediction capability enhanced by multiple-feature fusing technique}
    \label{fig:enter-label}
\end{figure}

Simultaneously, raw CSI data is represented in a machine-learning-friendly format. The RX positions are encoded using a positional encoding module, while the CSI values undergo an amplitude-phase decomposition to separate the magnitude and phase information. 

The final stage integrates the environmental features and CSI data to predict spatial CSI. Features extracted from the environment and encoded RX locations are again concatenated as tensors before being fed into the learning model.The tensors are saved for reuse by multiple models. Notice that feeding positions and the raw map separately into the model will cause in a degenerate effect of feature learning, due to that the features are not aligned.

Multiple machine learning models are employed to predict spatial CSI by leveraging knowledge of physical environments. The models utilized include GPT-2, VAE, Transformer, Diffusion Model which incorporates a U-Net inside, and MLP as baseline.

\subsection{Results and Discussion}
The models are all evaluated using NMSE metric. 

In this task, the VAE achieves the best performance, represented by a smooth and gradually decreasing NMSE curve in both training and validation sets, as shown in Fig.9. Meanwhile the time cost by VAE training is less than typical large models.

\begin{figure}
    \centering
    \includegraphics[width=0.8\linewidth]{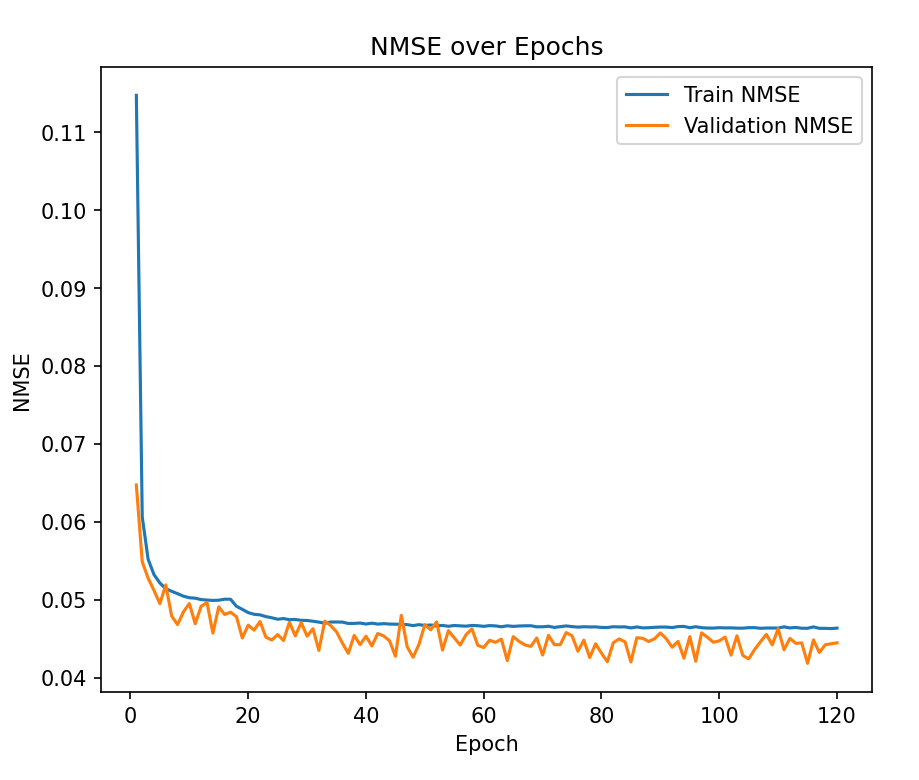}
    \caption{The learning curve of VAE, exhibiting high accuracy and low NMSE in the task of predicting spatial CSI distributed in an area of CUHK(SZ)}
    \label{fig:enter-label}
\end{figure}

The Diffusion Model also shows robust performance in prediction. During early epochs, the NMSE might fluctuate. However, as training progresses, the curve converges and stabilizes. Considering no obvious progress in reducing NMSE after 30 epochs, early-stop is adopted and the lowest NMSE of validation set is 0.125 (Fig.10).

\begin{figure}
    \centering
    \includegraphics[width=0.9\linewidth]{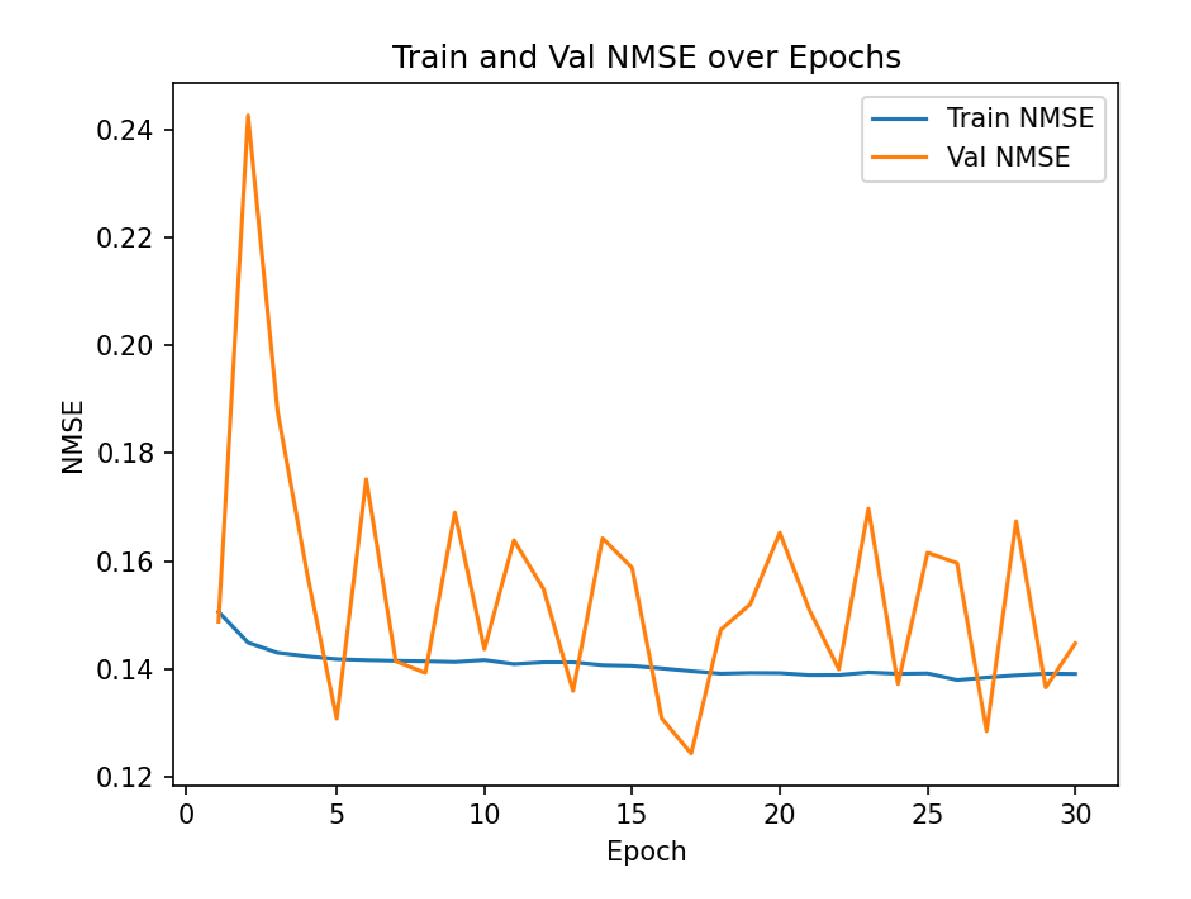}
    \caption{The learning cutve of Diffusion model}
    \label{fig:enter-label}
\end{figure}

GPT-2 and Transformer models, which are primarily tailored for sequence modeling, exhibit moderate initial improvement in NMSE, with the curve plateauing early. At the same time, The MLP model, being the simplest architecture, yields the highest NMSE across all epochs. The learning curves of all these three models decrease slowly and sporadically, reflecting their limited ability to capture CSI function. The validation NMSE of them are significantly higher than the training NMSE, indicating lack of fitting (Fig.11).
\begin{figure}
    \centering
    \includegraphics[width=0.9\linewidth]{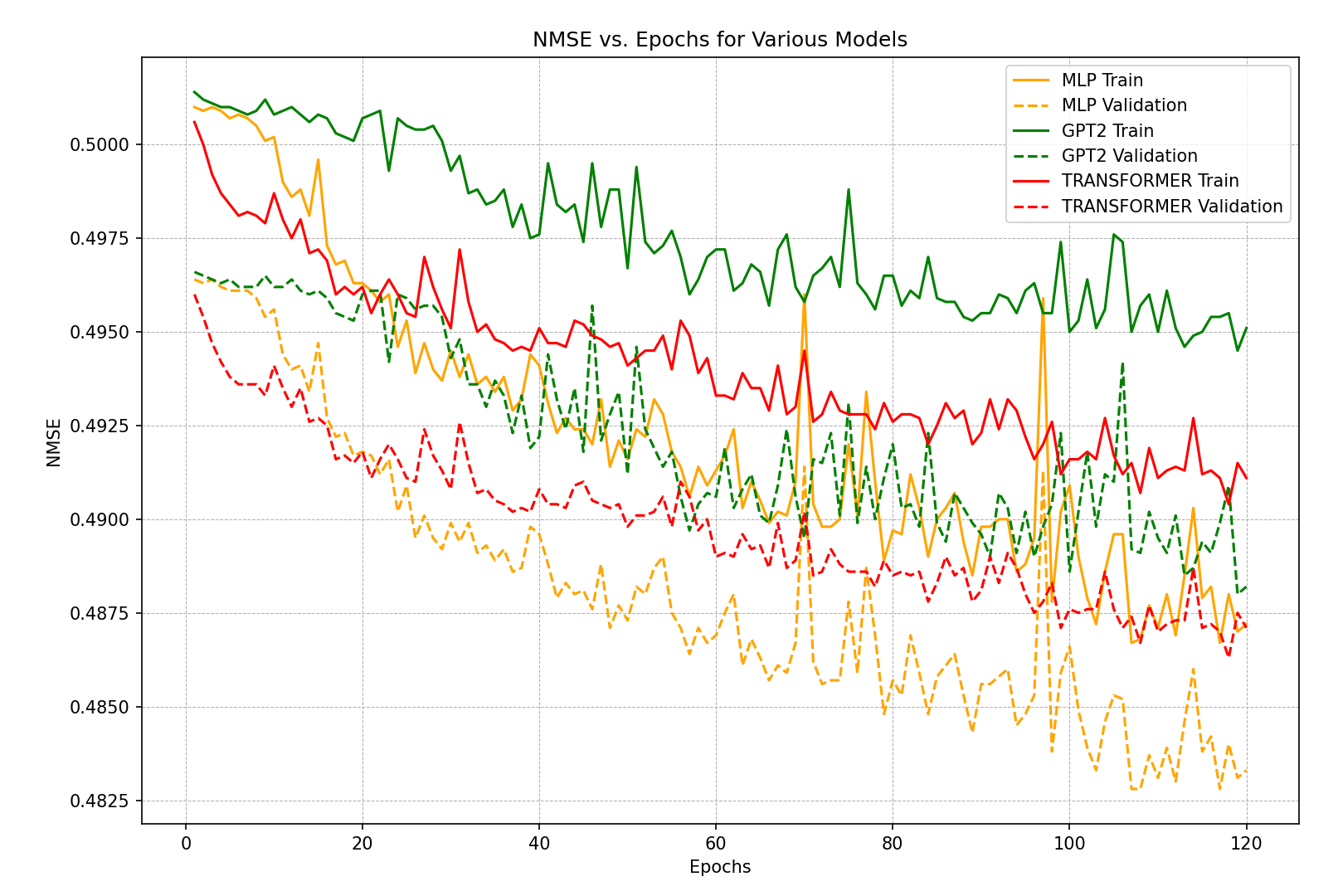}
    \caption{Learning curves of GPT-2, Transformer and MLP}
    \label{fig:enter-label}
\end{figure}

\begin{table*}[!h]
    \centering
    \caption{Training and Validation NMSE Across Models and Epochs}
    \begin{tabular}{lcccccccccccccc}
        \toprule
        Epoch & \multicolumn{2}{c}{Transformer} & \multicolumn{2}{c}{GPT-2} & \multicolumn{2}{c}{MLP} & \multicolumn{2}{c}{VAE} & \multicolumn{2}{c}{Diffusion} \\
        \cmidrule(lr){2-3} \cmidrule(lr){4-5} \cmidrule(lr){6-7} \cmidrule(lr){8-9} \cmidrule(lr){10-11}
        & Training & Validation & Training & Validation & Training & Validation & Training & Validation & Training & Validation \\
        \midrule
        1  & 0.5006 & 0.4960 & 0.5014 & 0.4966 & 0.5010 & 0.4964 & 0.1148 & 0.0647 & 0.1514 & 0.1496 \\
        10 & 0.4987 & 0.4941 & 0.5008 & 0.4962 & 0.5002 & 0.4956 & 0.0502 & 0.0495 & 0.1424 & 0.1445 \\
        20 & 0.4962 & 0.4918 & 0.5007 & 0.4960 & 0.4963 & 0.4918 & 0.0483 & 0.0467 & 0.1400 & 0.1661 \\
        30 & 0.4951 & 0.4908 & 0.4993 & 0.4943 & 0.4945 & 0.4899 & 0.0473 & 0.0453 & 0.1398 & 0.1456 \\
        60 & 0.4933 & 0.4890 & 0.4972 & 0.4906 & 0.4913 & 0.4869 & 0.0466 & 0.0439 &  &  \\
        90 & 0.4919 & 0.4871 & 0.4916 & 0.4876 & 0.4902 & 0.4866 & 0.0464 & 0.0457 &  &  \\
        120 & 0.4931 & 0.4863 & 0.4911 & 0.4871 & 0.4872 & 0.4833 & 0.0463 & 0.0445 &  &  \\
        \bottomrule
    \end{tabular}
    \label{tab:combined_nmse_values}
\end{table*}

\section{Conclusion}

In this paper, the study starts with probablistic modeling like QuaDRiGa, and switches to a more robust, physics-driven approach. By integrating raytracing-based channel generation technique, we successfully create a highly realistic dataset which reflects the sophisticated nature of real-world urban scenario. Combined with the dataset, the use of large models like GPT-2 and VAE provide an effective solution for predicting spatial CSI. Nevertheless, the incorporation of physical characteristics of the paths of the signal reduces the uncertainty of the function to be learned by the model, hence reliably raises the accuracy of prediction. The novel methodology is believed to be adaptive to totally different and complex urban maps.

\section{Future Developments and Possible Applications}

Future work on CSI prediction with machine learning could advance along multiple fronts. Adopting architectures such as Mistral AI or Llama models, or models capable of handling complex graphs is a promising direction. For the latter idea, applying GNN, VoxNet or PointNet could be helpful, but the challenge of the feature engineering also exists.

The application potential for CSI prediction is considerable. One of the most immediate applications lies in optimizing beamforming and antenna selection strategies in 5G and emerging 6G networks. Another promising application is in vehicular and mobile robotics communications, where vehicles or drones require robust and dynamic connection capabilities. In such cases, real-time CSI prediction enables adaptive signal transmission, leading to improved reliability in motion-intensive scenarios. In addition, the integration of these predictive models with other intelligent systems in smart cities or IoT environments offers potential for advancements in resource allocation, where predicted CSI data could inform better network resource deployment.


\begin{thebibliography}{99}

\bibitem{zhang2024}
L. Zhang, H. Sun, Y. Zeng, and R. Q. Hu, "Spatial Channel State Information Prediction with Generative AI: Towards Holographic Communication and Digital Radio Twin," arXiv:2401.08023, 2024.

\bibitem{al-asadi2023}
A. Al-Asadi, I. R. K. Al-Saedi, S. K. Alwane, H. Li, and L. Alzubaidi, "Enhanced MIMO CSI Estimation Using ACCPM with Limited Feedback," in Sensors, vol. 23, no. 18, p. 7965, Sep. 2023. [Online]. https://doi.org/10.3390/s23187965

\bibitem{gheryani2009}
M. Gheryani, "A new approach to the design of adaptive MIMO wireless communication systems," 2009. https://core.ac.uk/download/211515877.pdf

\bibitem{abd2018}
A. M. Abd El-Moaty and A. Zerguine, "Sparse Channel Estimation with Gradient-Based Algorithms: A comparative Study," 2018. https://arxiv.org/pdf/1812.04196

\bibitem{liu2019}
Z. Liu, L. Zhang, and Z. Ding, "Overcoming the Channel Estimation Barrier in Massive MIMO Communication Systems," 2019. https://arxiv.org/pdf/1912.10573

\bibitem{kim2020}
H. Kim, S. Kim, H. Lee, C. Jang et al., "Massive MIMO Channel Prediction: Kalman Filtering vs. Machine Learning," 2020. https://arxiv.org/pdf/2009.09967

\bibitem{pecorella2020}
T. Pecorella, R. Fantacci, and B. Picano, "Improving CSI Prediction Accuracy with Deep Echo State Networks in 5G Networks," 2020. https://www.ncbi.nlm.nih.gov/pmc/articles/PMC7697607/

\bibitem{zhang2022}
Y. Zhang, A. Alkhateeb, P. Madadi, J. Jeon et al., "Predicting Future CSI Feedback For Highly-Mobile Massive MIMO Systems," 2022. https://arxiv.org/pdf/2202.02492

\bibitem{ko2022}
B. Ko, H. Kim, and J. Choi, "Massive MIMO Channel Prediction Using Machine Learning: Power of Domain Transformation," 2022. https://arxiv.org/pdf/2208.04545

\bibitem{park2022}
S. Park and O. Simeone, "Speeding up Training of Linear Predictors for Multi-Antenna Frequency-Selective Channels via Meta-Learning," 2022. https://www.ncbi.nlm.nih.gov/pmc/articles/PMC9600732/

\bibitem{kim2022}
H. Kim, J. Choi, and D. J. Love, "Massive MIMO Channel Prediction Via Meta-Learning and Deep Denoising: Is a Small Dataset Enough?," 2022. https://arxiv.org/pdf/2210.08770

\bibitem{zhou2022}
T. Zhou, H. Zhang, B. Ai, C. Xue and L. Liu, "Deep-Learning-Based Spatial–Temporal Channel Prediction for Smart High-Speed Railway Communication Networks," in IEEE Transactions on Wireless Communications, vol. 21, no. 7, pp. 5333-5345, July 2022, doi: 10.1109/TWC.2021.3139384.  https://ieeexplore.ieee.org/document/9676455

\bibitem{du2023}
Y. Du, S. Chang Liew, K. Chen, and Y. Shao, "The Power of Large Language Models for Wireless Communication System Development: A Case Study on FPGA Platforms," 2023. https://arxiv.org/pdf/2307.07319

\bibitem{liu2024}
L. Liu, X. Liu, S. Gao, X. Cheng, and L. Yang, "LLM4CP: Adapting Large Language Models for Channel Prediction," arXiv:2406.14440, 20 Jun. 2024.

\bibitem{zhou2024}
H. Zhou, C. Hu, Y. Yuan, Y. Cui et al., "Large Language Model (LLM) for Telecommunications: A Comprehensive Survey on Principles, Key Techniques, and Opportunities," 2024. https://arxiv.org/pdf/2405.10825

\bibitem{winner2008}
Q. F. Channel Simulators for mmWave and 5G Applications. 2017. https://api.semanticscholar.org/CorpusID:56454171.

\bibitem{kyosti2008}
P. Kyösti, J. Meinilä, L. Hentilä, X. Zhao, T. Jämsä, C. Schneider, M. Narandžić, M. Milojević, A. Hong, J. Ylitalo, V.-M. Holappa, M. Alatossava, R. Bultitude, Y. de Jong, and T. Rautiainen, “WINNER II Channel Models,” 2008, https://www.cept.org/files/8339/winner - final report.pdf

\bibitem{FHI2023}
Fraunhofer Heinrich Hertz Institute, Wireless Communications and Networks, Quasi Deterministic Radio Channel Generator User Manual and Documentation, Document Revision: v2.8.1, Einsteinufer 37, 10587 Berlin, Germany, December 13, 2023.

\bibitem{sanchez}
Sánchez, D., Trujillo, M., Varela, P., Rattaro, C., Inglés, L., \& Belzarena, P. MIMO Simulation in 5G Networks: Py5cheSim and DeepMIMO Integration. In 2023 XLIX Latin American Computer Conference (CLEI) (pp. 1-7). 2023
\end{thebibliography}

\end{document}